\documentclass{article}

\usepackage{graphicx}
\usepackage{caption}
\usepackage{subcaption}
\usepackage{microtype}
\usepackage{amsmath}
\usepackage{booktabs} 
\usepackage{color}

\usepackage{hyperref}

\usepackage{multirow}
\usepackage{bbding}
\usepackage{colortbl}
\usepackage{soul}

\usepackage{diagbox}

\definecolor{Gray}{gray}{0.9}
\sethlcolor{Gray}

\newcommand{\W}{\mathbf{W}}
\newcommand{\B}{\mathbf{B}}

\newcommand{\X}{\mathbf{X}}

\newcommand{\Y}{\mathbf{Y}}

\usepackage[ruled,linesnumbered]{algorithm2e}
\usepackage{amssymb}
\usepackage[accepted]{icml2023}

\icmltitlerunning{Bi-directional Masks for Efficient N:M Sparse Training}

\begin{document}

\twocolumn[

%
\icmltitle{Bi-directional Masks for Efficient N:M Sparse Training}


\icmlsetsymbol{equal}{*}

\begin{icmlauthorlist}
\icmlauthor{Yuxin Zhang*}{mac}
\icmlauthor{Yiting Luo*}{mac}
\icmlauthor{Mingbao Lin}{yt}
\icmlauthor{Yunshan Zhong}{mac}
\icmlauthor{Jingjing Xie}{mac}
\icmlauthor{Fei Chao}{mac}
\icmlauthor{Rongrong Ji}{mac,xmuai,pcl}
\end{icmlauthorlist}

\icmlaffiliation{mac}{Media Analytics and Computing Laboratory, Department of Artificial Intelligence, School of Informatics, Xiamen University, Xiamen, China}
\icmlaffiliation{yt}{Tencent Youtu Lab, Shanghai, China}
\icmlaffiliation{xmuai}{Institute of Artificial Intelligence, Xiamen University, Xiamen, China}
\icmlaffiliation{pcl}{Pengcheng Lab, Shenzhen, China}
\icmlcorrespondingauthor{Rongrong Ji}{rrji@xmu.edu.cn}

\icmlkeywords{Machine Learning, ICML}

\vskip 0.3in
]



\printAffiliationsAndNotice{\icmlEqualContribution} 
\begin{abstract}
We focus on addressing the dense backward propagation issue for training efficiency of N:M fine-grained sparsity that preserves at most N out of M consecutive weights and achieves practical speedups supported by the N:M sparse tensor core.
Therefore, we present a novel method of Bi-directional Masks (Bi-Mask) with its two central innovations in:
1) Separate sparse masks in the two directions of forward and backward propagation to obtain training acceleration. It disentangles the forward and backward weight sparsity and overcomes the very dense gradient computation.
2) An efficient weight row permutation method to maintain performance. It picks up the permutation candidate with the most eligible N:M weight blocks in the backward to minimize the gradient gap between traditional uni-directional masks and our bi-directional masks.
Compared with existing uni-directional scenario that applies a transposable mask and enables backward acceleration, our Bi-Mask is experimentally demonstrated to be more superior in performance. Also, our Bi-Mask performs on par with or even better than methods that fail to achieve backward acceleration.
Project of this paper is available at \url{https://github.com/zyxxmu/Bi-Mask}.

\end{abstract}

\section{Introduction}\label{sec:introduction}
%
%
%
%

The past decade has witnessed thriving deep neural networks (DNNs) in various machine learning applications~\cite{he2016deep,he2017mask,girshick2014rich}. In large part, the prosperity is driven by increasing parameters and computations, which however, make DNN models too cumbersome to be deployed on resource-constrained edge devices such as cell phones and Internet-of-Things (IoT) devices. 
Therefore, the research community is sorely in need of technical renovation to compress the DNNs~\cite{hubara2016binarized, tan2019efficientnet, lin2020hrank}.

\begin{figure}[!t]
\centering
\begin{subfigure}{1\linewidth}
\centering
\includegraphics[width=0.98\linewidth]{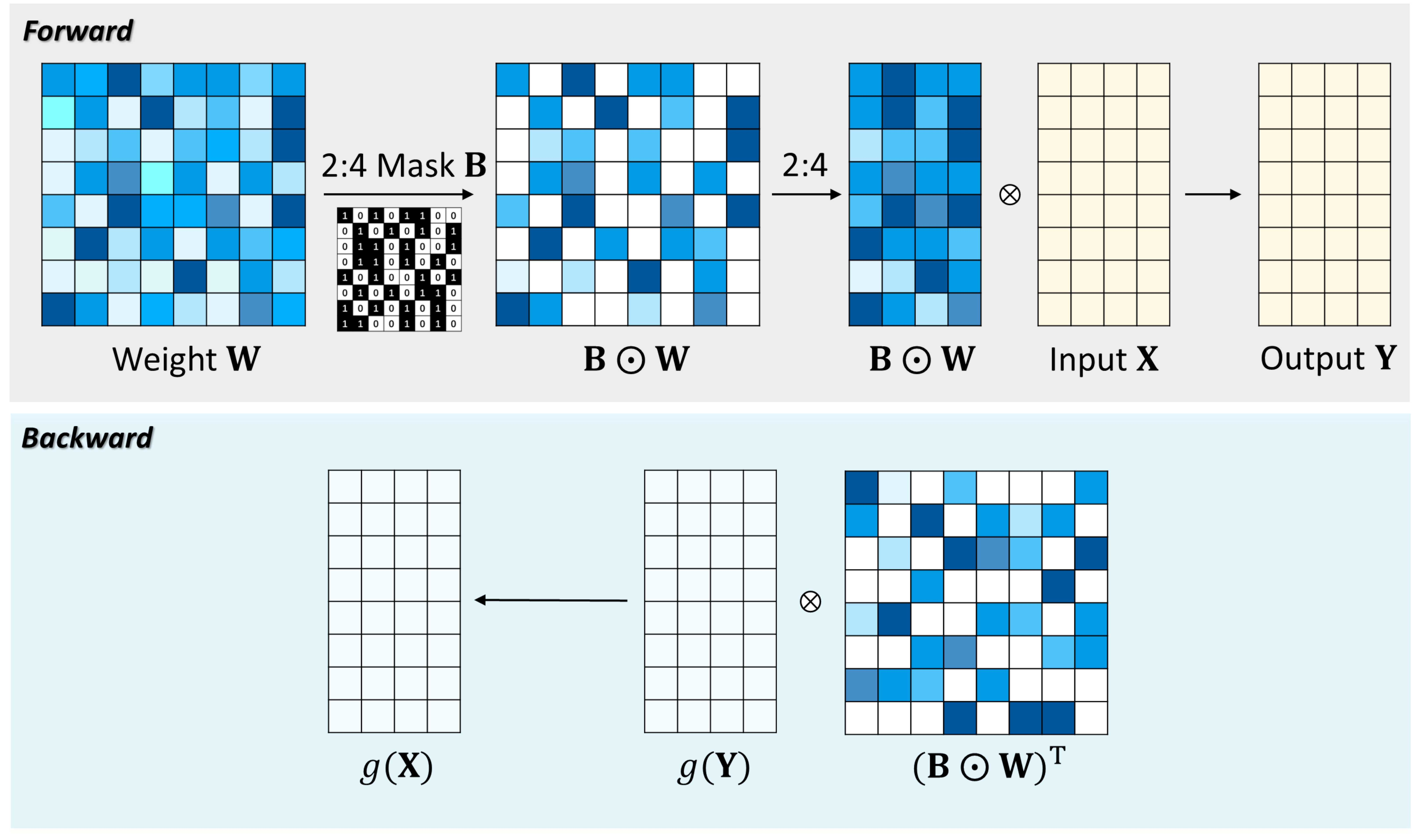}
\caption{Vanilla N:M Mask\label{fig:vanilla}}
\label{fig1a}
 \end{subfigure}

 \begin{subfigure}{1\linewidth}
\centering
\includegraphics[width=0.98\linewidth]{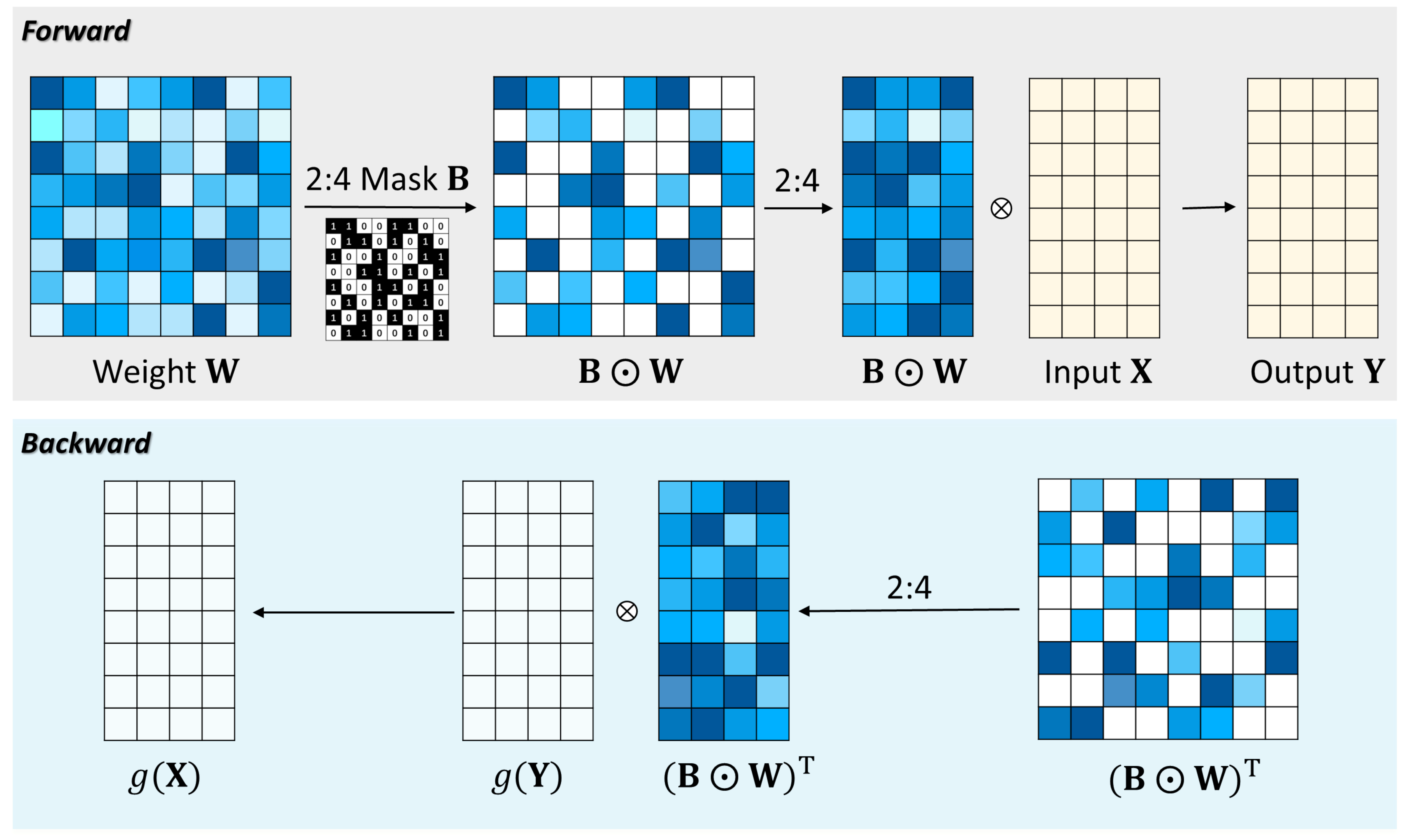}
\caption{Transposable N:M Mask\label{fig:transposable}}
\label{fig1b}
 \end{subfigure}
 \caption{Comparison between vanilla N:M mask and transposable N:M mask (2:4 case). The vanilla N:M mask~\cite{zhou2021learning,nvidia2020a100} generates sparse weights with N:M property in rows, leading to forward acceleration but remaining dense backward propagation as the weight transposition operation impairs N:M blocks. The transposable N:M mask~\cite{hubara2021accelerated} generates sparse weights that have N:M property in both rows and columns, leading to forward \& backward acceleration. Both methods consider only one sparse mask.}
\vspace{-0.2cm}
\end{figure}

%
%
%
%
%
%

\begin{figure*}[!t]
    \centering
    \includegraphics[width=1\textwidth]{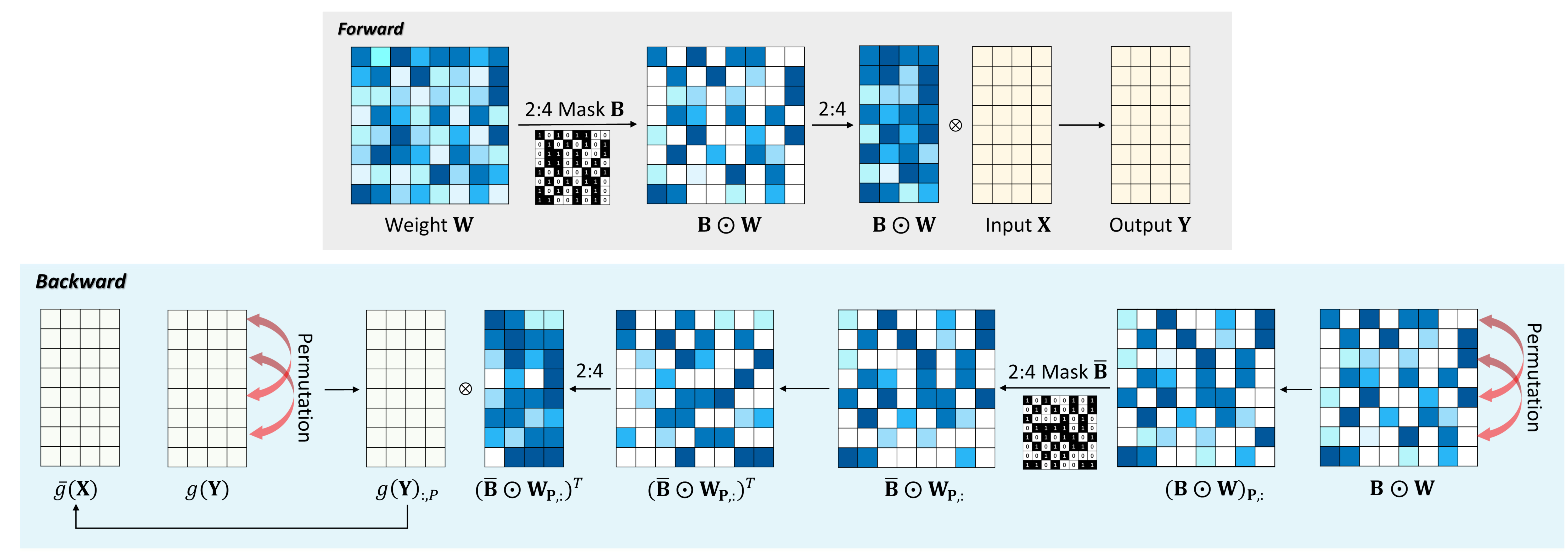}
    \caption{Framework of the proposed Bi-direction Masks (Bi-Mask). It separately builds two N:M sparse masks in the forward and backward direction, thus enabling training acceleration in both directions. During backward propagation, Bi-Mask performs an efficient row permutation to make the sparse weights have more eligible N:M weight blocks before generating the backward mask.}
    \label{fig:framework}
\end{figure*}

By removing redundant network weights~\cite{lecun1989optimal, han2015learning, he2017channel}, network sparsity has emerged as a piece of modern equipment to obtain a lightweight sparse model. 
Through removing individual weights at arbitrary positions, fine-grained sparsity is demonstrated to reach a high sparse ratio with performance guarantee~\cite{han2015learning, evci2020rigging}. Unfortunately, the resulting unstructured sparse weights hardly produce acceleration on off-the-shelf hardware.
Coarse-grained sparsity is more hardware friendly as it typically removes an entire weight block~\cite{ji2018tetris, meng2020pruning} or convolution filter~\cite{liu2019metapruning, lin2020hrank}.
In comparison with fine-grained sparsity, the compressed model gains noticeable speedup, yet suffers more performance degradation.
Therefore, it is a challenging yet valuable issue to simultaneously retain model performance of DNN models and achieve hardware acceleration.

%
%
%
%
%

Luckily, recent N:M fine-grained sparsity has provided a promising solution. By requiring at most N non-zero elements out of every M contiguous weights, N:M sparsity includes the performance advantage of fine-grained sparsity as well as practical acceleration thanks to the hardware innovation of N:M sparse tensor core~\cite{ronny2020nvidia,fang2022algorithm}.
Nvidia~\citep{nvidia2020a100} has presented the ASP (APEX’s Automatic Sparsity) paradigm that achieves 2:4 sparsity within three steps, unfolded as training a dense network, applying 2:4 fine-grained sparsity using magnitude-based pruning~\cite{han2015learning}, and re-training the sparse network.
Despite the satisfying performance, ASP exhibits drawbacks in its tedious training cost as it contains dense network training and N:M sparse retraining.
This largely prohibits the application of N:M sparsity when confronting with scarce training resources.
%

%
%
%
%
%
%

The above issue has been partially addressed by directly training an N:M sparse network from scratch~\cite{zhou2021learning}. 
Yet, the sparse tensor core is only utilized to accelerate the forward multiplication during training. As illustrated in Fig.\,\ref{fig:vanilla}, the weight transposition operation in the backward impairs N:M blocks and thus fails to support acceleration in gradient calculation.
To mitigate this, ~\cite{hubara2021accelerated} proposed N:M transposable mask, where a binary mask that indicates whether to remove weights is required to have N:M property along the rows and columns. Therefore, after performing transposition, it still satisfies the N:M format as shown in Fig.\,\ref{fig:transposable}.
Unfortunately, the transposable requirement is observed to have more performance degradation, which is presumably caused by less flexibility of the sparsity pattern~\cite{hubara2021accelerated}.
In Sec.\,\ref{trans}, we further show severe performance degradation at a higher sparse level such as 1:8 and 1:16.
We therefore reflect on this: how can we address the efficiency of N:M sparse training without a compromise on performance?

In this paper, we attempt to answer the above question by introducing a novel method of Bi-directional Masks (Bi-Mask) that performs surprisingly well without any N:M transposable constraint. Fig.\,\ref{fig:framework} illustrates framework of our Bi-Mask. In particular, along the forward and backward directions, two separate binary masks are constructed according to the weight magnitude~\cite{han2015learning}. As a contrast, we require the forward mask to follow N:M property in its rows while in columns for the backward mask. By this way, we concurrently enable forward \& backward acceleration from the N:M sparse tensor core. Also, the bi-directional masks benefit performance from more flexible sparsity pattern.
Nevertheless, they also bring about deficiency of gradient gap since the backward mask modifies the gradient of forward loss.
Given this issue, an efficient row permutation is further introduced before enforcing the backward mask.
In detail, we first change row order of weight matrix and then pick up the permutation with the most eligible N:M weight blocks from a dozen of candidates. By changing column order of output gradient accordingly, we succeed in retaining the same outputs between with/without row permutation, and at the same time well reducing the gradient gap between uni-directional/bi-directional mask(s).

\begin{table}[!t]
    \caption{Advantage comparison between the vanilla N:M mask (Mask)~\cite{nvidia2020a100, zhou2021learning}, the transposable N:M mask (T-Mask)~\cite{hubara2021accelerated} and our proposed Bi-direction Mask (Bi-Mask) for N:M sparse training.
    }
    \centering
\resizebox{1.0\columnwidth}{!}{
    \begin{tabular}{cccc>{\columncolor[gray]{0.9}}c}
    \toprule
  Advantage &   Vanilla Mask  &  T-Mask &   Bi-Mask\\
   \midrule
  Forward Acceleration  & \Checkmark & \Checkmark & \Checkmark \\
 Backward Acceleration& \XSolidBrush  &\Checkmark & \Checkmark \\
 Performance Maintenance&   \Checkmark &\XSolidBrush  & \Checkmark\\
    \bottomrule
    \end{tabular}}
    \label{tab:merit_comparison}
\end{table}

Our simple design of Bi-Mask turns out to achieve remarkable results. 
Besides forward \& backward training acceleration, Bi-Mask well improves the performance of transposable mask (T-Mask) across different N:M patterns, benchmarks, and networks.
For example, Bi-Mask achieves 71.5\% Top-1 accuracy when training 1:16 sparse ResNet-50 on ImageNet, surpassing T-mask by 5.3\%.
More surprisingly, our approach achieves comparable or even better results than vanilla N:M methods, where the backward propagation can not be accelerated. 
For example, our Bi-Mask exceeds Top-1 accuracy of SR-STE~\cite{zhou2021learning} by 0.4\% when training 2:4 sparse ResNet-50 on ImageNet.
Table\,\ref{tab:merit_comparison} provides advantage comparison between different mask methods.

\section{Related Work}\label{sec:related_work}

\subsection{Network Sparsity}
Network sparsity has been one of the most effective tools to relieve the complexity of DNNs over the past decades~\cite{lecun1989optimal, han2015learning}. Pioneering works implement network sparsity in a fine-grained granularity where weights at arbitrary positions are removed to obtain a compact network.
~\cite{han2015learning} presented a classic three-step paradigm including pre-training a full network, removing low-magnitude weights, and fine-tuning the sparse networks.
The lottery ticker hypothesis~\cite{frankle2018lottery} further reveals the existence of randomly-initialized sparse networks that can be trained independently to compete with the performance of the dense model.
In principle, the fine-grained network sparsity can maintain the performance of dense networks at an ultra-high sparse ratio like 90.0\%~\cite{mostafa2019parameter, blalock2020state}. 
Nevertheless, it receives very limited speedups since the resulting sparse networks are in unstructured formats, which barely take advantage of general hardware platforms.
%

%
%
%
%
%
%

%
Coarse-grained sparsity targets at removing entire weight blocks~\cite{ji2018tetris, meng2020pruning} or convolution filters~\cite{liu2018rethinking, lin2020hrank} to make the sparse networks compatible with off-the-shelf hardware.
For instance,~\cite{li2016pruning} removed convolution filters with smaller $\ell_1$ norm, while~\cite{lin2020hrank} considered the rank of feature maps as the filter importance measurement.
Unfortunately, coarse-grained sparsity suffers severe performance drops at sparsity levels higher than 50\% due to the flexibility constraint on network sparsity~\cite{renda2020comparing}.
Different from the existing sparsity granularity, this paper focuses on N:M fine-grained sparsity~\cite{zhou2021learning, sun2021dominosearch, pool2021channel}, which preserves at most N out of M consecutive weights. In addition to performance maintenance, N:M sparsity is also able to obtain practical acceleration from the hardware innovation of N:M sparse tensor core~\cite{nvidia2020a100, fang2022algorithm}.

%
%
%

\subsection{Sparse Training}
Sparse training serves as an effective tool to boost the performance of network sparsity~\cite{hoefler2021sparsity,evci2020rigging,sanh2020movement}.
It dynamically prunes and revives weights of the sparse networks during training according to specific criteria.
%
For example, RigL~\cite{evci2020rigging} removes smaller-magnitude weights and revives weights with larger-magnitude gradients.
%
Besides, sparse momentum~\cite{dettmers2019sparse} considers magnitude of mean weight momentum as a guide to redistribute the sparse weights.
%
%
%
In this paper, we focus on training N:M sparse networks.
As a study mostly related to ours, the transposable N:M masks~\cite{hubara2021accelerated} requires one single sparse mask with N:M blocks in both rows and columns such that the transposition in the backward also embraces hardware acceleration.
In contrast, our method separately builds sparse masks in the forward and backward propagation without additional sparse constraints and gains significantly better performance under the same N:M case.
Besides,~\cite{pool2021channel} proposed to permute the input channel of pre-trained CNNs to maximally preserve the magnitude of N:M sparse networks.
Very differently, our Bi-Mask permutes the row dimension of sparse weights that are trained from scratch, with a diverse object of obtaining more eligible N:M weight blocks to mitigate the gradient gap in the backward sparsity.

\section{Methodology}\label{sec:method}


\subsection{Revisiting N:M Sparse Training}

%
%
%
%
%
%
%

We first introduce some basic knowledge about the N:M fine-grained sparsity. Let $\W \in \mathbb{R}^{I \times J}$ be the parameter matrix from a specific network layer.
Considering the input tensor $\X$, the forward propagation represented with form of matrix multiplication can be formulated as:
\begin{equation}\label{eq:forward}
    \Y = \W * \X,
\end{equation}
where $\Y$ is the output tensor and $*$ is the matrix multiplication operation.
N:M sparsity forces at most N out of M consecutive weights in the rows of $\W$ to have non-zero values. 
The sparsity can be achieved via a binary matrix $\B \in \{0,1\}^{I\times J}$ where a block of every M contiguous elements contains at most N as:
\begin{equation}\label{eq:contraint_forward}
{\Vert\B_{i, j:j+\text{M}}\Vert}_0 \le \text{N},
\end{equation}
in which $i = 1, 2, 3, ..., I$ and $j =1, \text{M}, 2\text{M}, ..., J $. 
Then, the sparse forward propagation can be formulated as:
\begin{equation}\label{eq:sparse_forward}
    \Y = (\B \odot \W) * \X,
\end{equation}
where $\odot$ denotes the element-wise multiplication. Since $\B \odot \W$ meets N:M requirement, the matrix multiplication with $\X$ can be efficiently implemented by the N:M sparse tensor core, as illustrated in Fig.\,\ref{fig:vanilla}.
%

%
%
%
%
%
%

%
N:M sparse training starts from randomly-initialized networks~\cite{zhou2021learning,zhang2022learning}, thus avoiding heavy burden of pre-training a dense model~\cite{nvidia2020a100}.
We base our study on the popular SR-STE ~\cite{zhou2021learning} for N:M sparse training, simply illustrated for ease of understanding in the following.
During forward propagation, it adapts the binary mask $\B$ at each iteration as:
\begin{equation}\label{eq:weight_projection}
\B_{i, j+m} = \left\{ \begin{array}{ll} 
 0, \; \textrm{if} \; |\W_{i, j+m}| < \text{Top}(|\W_{i, j:j+\text{M}}|, \text{N}),\\
 1, \; \text{otherwise},
  \end{array} \right.
\end{equation}
where $1 \leq m \leq \text{M}$, $|\cdot|$ represents the absolute function, and $\text{Top}(|\W_{i, j:j+\text{M}}|, \text{N})$ returns the N-$th$ largest value within $|\W_{i, j:j+\text{M}}|$. 
Therefore, we obtain the forward binary mask according to the weight magnitude in each block.
During backward propagation, the gradients of $\B \odot \W$ are directly passed to $\W$ according to the straight-through-estimator (STE)~\cite{bengio2013estimating}.
%


\subsection{Rethinking the Transposable N:M Mask}\label{trans}

The above sparse mask is indeed uni-directional towards forward propagation. By forming N:M blocks in rows of the mask, Eq.\,(\ref{eq:sparse_forward}) permits forward acceleration from the N:M sparse tensor core between the weights and inputs. Unfortunately, such a vanilla mask crashes backward acceleration due simply to the transposition operation. To explain, the gradient in the backward propagation is computed as:
\begin{equation}\label{eq:backward}
    g(\X) = (\B \odot \W)^T * g(\Y),
\end{equation}
where $g(\cdot)$ denotes the gradient with respect to its input. The above equation requires $(\B \odot \W)^T$ to have N:M blocks in rows for accelerating multiplication with $g(\Y)$, however, it is in columns on account of the transposition operation. Thus, the backward propagation remains dense and fails to be accelerated, as illustrated in Fig.\,\ref{fig:vanilla}.

To address this issue, ~\cite{hubara2021accelerated} presented transposable N:M mask that is required to satisfy row-wise and column-wise N:M blocks such that the transposition also undertakes an important mission of N:M property in rows.
Consequently, the binary mask $\B$ is constrained as:
\begin{equation}\label{eq:constraint_transposable}
{\Vert\B_{i, j:j+\text{M}}\Vert}_0 \le \text{N}, 
 \quad {\Vert\B_{k:k+\text{M}, l}\Vert}_0 \le \text{N}, 
\end{equation}
where $i = 1, 2, 3, ..., I$, $j = 1, \text{M}, 2\text{M}, ..., J$, $k = 1, M, 2M, ..., I$, and $l = 1, 2, 3, ..., J$.
Besides, ~\cite{hubara2021accelerated} further introduced a $2$-approximation algorithm to reduce complexity of finding the transposable mask.
%

%
\begin{figure}[!t]
    \centering
    \begin{subfigure}{1\linewidth}
    \centering
    \includegraphics[width=0.98\linewidth]{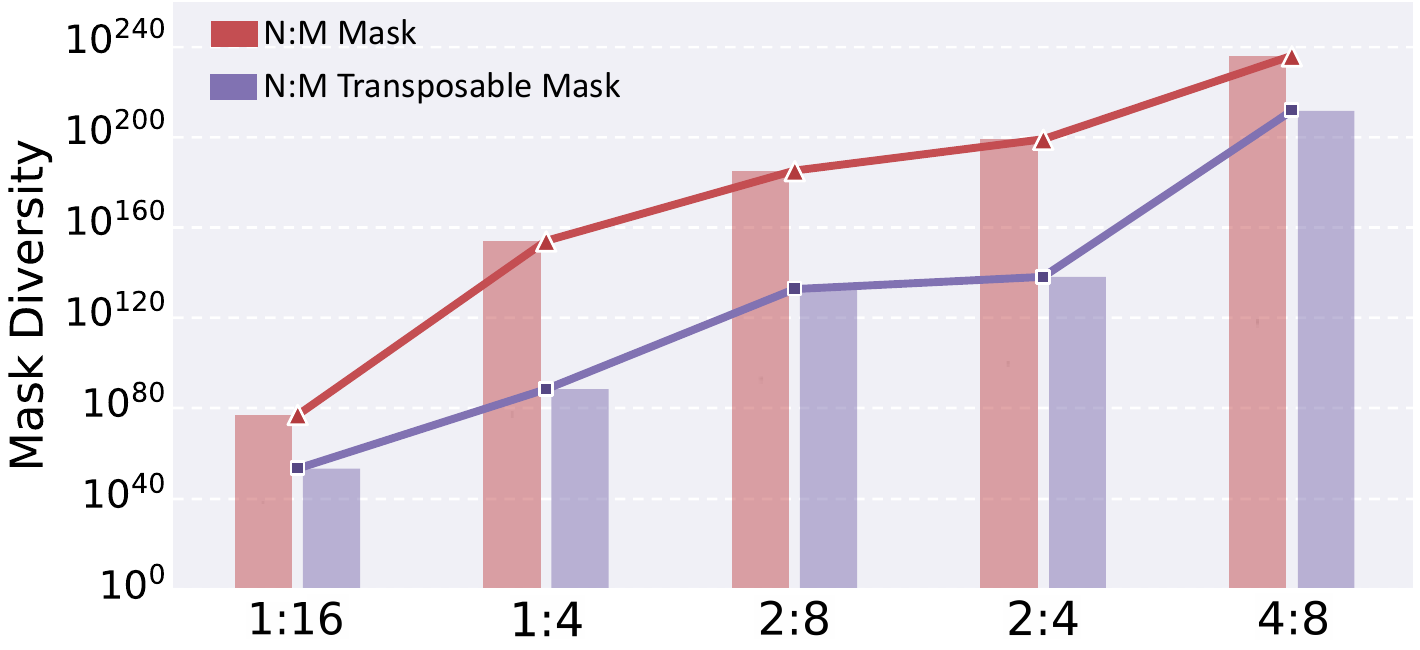}
    \caption{Flexibility Comparison\label{fig:mask-diversity}}
    \label{fig2a}
    \end{subfigure}

    \begin{subfigure}{1\linewidth}
    \centering
    \includegraphics[width=0.98\linewidth]{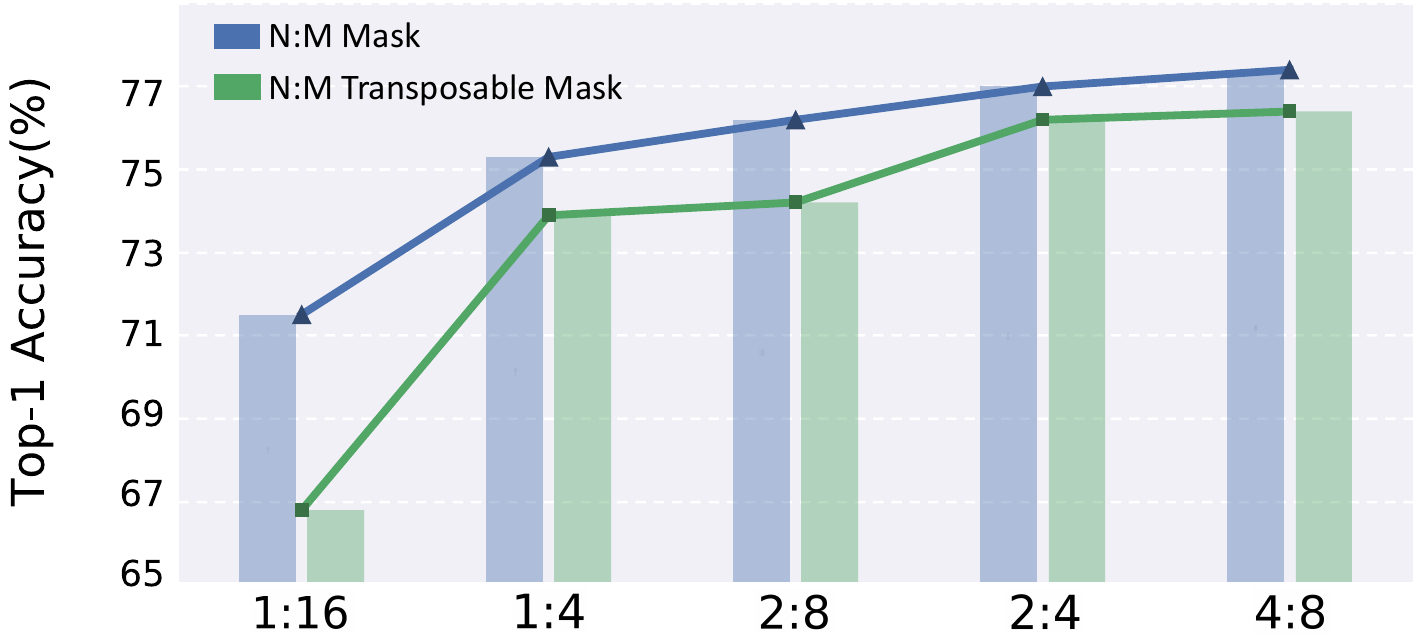}
    \caption{Performance Comparison\label{fig:performance}}
    \label{fig2b}
    \end{subfigure}
    \vspace{-0.2cm}
    \caption{Comparison between vanilla mask and transposable mask including (a) flexibility measured by mask diversity~\cite{hubara2021accelerated} and (b) performance of training sparse ResNet-50~\cite{he2016deep} on ImageNet~\cite{deng2009imagenet}.
    }
    \label{fig:mask_diversity}
    \vspace{-0.cm}
\end{figure}


%
%
%
%
%
%
%

Here we rethink the transposable pursuit for N:M sparse training. Although it enables backward acceleration, the flexibility of sparse networks is greatly restricted, which comes at the cost of performance degradation.
We first report the flexibility comparison between vanilla mask and transposible mask under different N:M cases. Fig.\,\ref{fig:mask-diversity} measures the flexibility using mask diversity that calculates the number of all possible masks under a given N:M mask case~\cite{hubara2021accelerated}. We can see a drastic flexibility degradation, in particular in cases of a small N or M.
As a consensus~\cite{gale2019state, nvidia2020a100}, more restrictions on sparse patterns incur worse performance of sparse networks. 
For example, unstructured sparsity~\cite{han2015learning} that removes arbitrary weights generally performs much better than structured sparsity~\cite{li2016pruning} that removes entire filter weights.
Consequently, severe performance occurs in transposable mask in comparison with the vanilla method, as we experimentally verify in Fig.\,\ref{fig:performance}, notably very poor 1:8 and 1:16.
The uni-directional masks, either vanilla or transposable, do not accomplish N:M backward acceleration without a compromise on performance. 
Therefore, in what follows, we address this issue from the perspective of bi-directional masks.

%
%
%

\subsection{Bi-directional N:M Masks}\label{sec:bi-mask}

In this section, we formally present our Bi-directional N:M masks (Bi-Mask). As its name suggests, our Bi-Mask disentangles the forward \& backward weight sparsity by involving two different masks during N:M sparse training.
Concretely speaking, in the forward direction, we count on the vanilla N:M mask $\B$ from Eq.\,(\ref{eq:contraint_forward}) that calls for N:M in rows to ensure the forward acceleration and results in better flexibility than the transposable N:M mask as we report in Fig.\,\ref{fig:mask-diversity}.
Very differently, we additionally build another mask $\bar{\B} \in \{0, 1\}^{I \times J}$ in the backward direction with the N:M requirement on its columns as:
\begin{equation}
    {\Vert\bar{\B}_{k:k+\text{M}, l}\Vert}_0 \le \text{N}, 
\end{equation}
in which $k=1, \text{M}, 2\text{M}, ..., I$, and $l = 1, 2, 3,..., J$. In this fashion, the backward acceleration is supported as well without a compromise on the flexibility of backward mask, and the backward gradient $g(\X)$ in Eq.\,(\ref{eq:backward}) is represented by the following approximation:
\begin{equation}\label{eq:gradient_calculation}
    \bar{g}(\X) =  (\bar{\B} \odot \W)^T * g(\Y).
    \vspace{-0.2cm}
\end{equation}

Nevertheless, the forward $\B$ requires gradient of $g(\X)$ for our Bi-Mask, which yields a gradient gap between practical bi-directional gradient $\bar{g}(\X)$ and ideal uni-directional gradient $g(\X)$. 
To solve this issue, we adapt the backward mask $\bar{\B}$ to the magnitudes of masked weights during sparse training as follows:
%
%
%
%
%
\begin{equation}\label{eq:backward_mask}
\bar{\B}_{k+m, l} = \left\{ \begin{array}{ll} 
 0, \; \textrm{if} \; |(\B \odot \W)_{k+m, l}| \\ \qquad < \text{Top}(|(\B \odot \W)_{k:k+\text{M}, l}|, \text{N}),\\
\B_{k+m, l}, \;  \text{otherwise},
  \end{array} \right.
\end{equation}
where $k = 1, \text{M}, 2\text{M}, ..., I$, $l = 1, 2, ..., J$, and $1 \le k \le \text{M}$.
For a deeper analysis, it is easy to understand that $\B_{k+m, l} = 0$ is a fully not necessary condition of $\B_{k+m, l} = 0$. That is, the event $\B_{k+m, l} = 0$ will produce the event $\B_{k+m, l} = 0$, but is not the only way for $\B_{k+m, l} = 0$ to occur.

The rationale behind Eq.\,(\ref{eq:backward_mask}) is two-fold: 
1) It maximizes the similarity of forward and backward masks by setting $\bar{\B}_{k+m, l} = \B_{k+m, l}$ if the magnitude of $\W_{k+m,l}$ is beyond the top-N largest. 
2) Applying our backward mask does not affect the updating of these weights with zero forward masks since $\B_{k+m, l} = 0$ always results in $\bar{\B}_{k+m, l} = 0$.
%
%
%
Unfortunately, it is a possibility that $\bar{\B}_{k+m, l} = 0$ does not necessarily result from $\B_{k+m, l} = 0$, in which case gradients of some non-zero masked weights are mistakenly eliminated, incurring performance degradation.

\begin{algorithm}[!t]
\SetKwInOut{Input}{Require}
\SetKwInOut{Output}{Output}
\SetAlgoLined
    \caption{Bi-Mask for Efficient N:M Sparse Training.}
    \label{alg:bimask}
    \Input{Iteration interval $\Delta T$, permutation candidate number $K$, weight matrix $\W$, training iteration $T$.}
    \Output{ Trained sparse weights $\W \odot \B$.}
    \For{$t \in [1, 2, \dots, T]$}{
        
        Obtain the forward mask $\B$ \text{via} Eq.\,(\ref{eq:weight_projection}); 
        
        Forward propagation \text{via} Eq.\,(\ref{eq:sparse_forward}); 
        
        \If {$t \; \% \; \Delta T = 0$} 
        {
        Randomly generate $K$ permutations and pick up the one as $P$ with the most eligible N:M blocks; 
        }
        
        Obtain the backward mask $\bar{\B}$ \text{via} Eq.\,(\ref{eq:backward_mask_final}); 
        
        Backward propagation \text{via} Eq.\,(\ref{eq:backward_final}); 
        
        Update via the SGD optimizer;
    }
    
    Return $\W \odot \B$. 
\end{algorithm}

To decrease this possibility, we continue a row permutation method along the row dimension of $\B \odot \W$.
Our major motivations are also two-fold:
1) We can see from Eq.\,(\ref{eq:backward_mask}) that the resulting mask block $\bar{\B}_{k:k+\text{M}, l}$ would exactly match with ${\B}_{k:k+\text{M}, l}$ if $(\B \odot \W)_{k:k+\text{M}, l}$ has N:M sparsity, and no gradient gap would occur.
2) Performing row permutation of $\B \odot \W$ improves the number of eligible N:M blocks as illustrated in Fig.\,\ref{fig:framework}. Importantly, it does not violate the gradient computation. Denoting $P \in \mathbb{N}^{I}$ as a permutation of $\{1,2,3,...,I\}$, the backward gradient $\bar{g}(\X)$ in Eq.\,(\ref{eq:gradient_calculation}) can be equally computed as:
\begin{equation}\label{eq:backward_final}
    \bar{g}(\X) =  \big(\bar{\B} \odot  (\W_{P, :})\big)^T * \big(g(\Y)_{:, P}\big),
\end{equation}
where the backward mask $\bar{\B}$ is computed based on the permutated $(\B \odot \W)_{P, :}$ accordingly:
\begin{equation}\label{eq:backward_mask_final}
\bar{\B}_{k+m, l} = \left\{ \begin{array}{ll} 
 0, \; \textrm{if} \; \Big|\big((\B \odot \W)_{P, :}\big)_{k+m, l}\Big| \\ \quad \quad \; < \text{Top}(\Big|\big((\B \odot \W)_{P, :}\big)_{k:k+\text{M}, l}\Big|, \text{N}),\\
(\B_{P,:})_{k+m, l}, \;  \text{otherwise}.
  \end{array} \right.
\end{equation}

Therefore, we only need to find a permutation $P$ that results in more eligible N:M blocks in each column of $(\B \odot \W)_{P,:}$. More N:M blocks decrease the possibility of eliminating gradients of non-zero masked weights. 
To avoid the cumbersome $I!$ possible permutations at each training iteration, we update a good permutation at a regularly spaced interval of every $\Delta T$ training iterations, and at each interval pick up the one that leads to the most eligible N:M blocks from randomly generating $K$ permutation candidates.

In Sec.\,\ref{sec:ablation}, we analyze that the permutation candidate number $K=100$ already returns good performance. Compared with the aforementioned $2$-approximation algorithm for the transposable N:M mask~\cite{hubara2021accelerated}, our method brings negligible runtime burden as we experimentally reported in Sec.\,\ref{imagenet}. Our algorithm presented in this paper is outlined in Alg.\,\ref{alg:bimask}.

\section{Experimentation}\label{sec:experiment}
\subsection{Settings}
\textbf{Datasets and Backbones.}
We conduct experiments on representative benchmarks for image classification. 
For small-scale dataset, we choose the CIFAR-10 dataset~\cite{krizhevsky2009learning}, which contains 60,000 32$\times$32 color images from 10 different classes, with 6,000 images for each class. 
For large-scale dataset, we choose the challenging ImageNet~\cite{deng2009imagenet}, which contains over 1.2 million images for training and 50,000 validation images in 1,000 categories. 
On CIFAR-10, we train N:M sparse ResNet-32~\cite{he2016deep} and MobileNet-V2~\cite{sandler2018mobilenetv2}.
On ImageNet, we train N:M sparse ResNet-18/50~\cite{he2016deep} and DeiT-small~\cite{touvron2021training}.
We compare our Bi-Mask with classic N:M sparse training methods including ASP~\cite{nvidia2020a100} and SR-STE~\cite{zhou2021learning} that fail backward acceleration, and transposable N:M mask (T-Mask)~\cite{hubara2021accelerated} that has backward acceleration.
Top-1 classification accuracy is reported for comparison on both datasets.

\textbf{Implementation Details.}
Our implementation of Bi-Mask is based on the PyTorch framework~\cite{pytorch2015}.
All experiments are conducted on the NVIDIA Tesla A100 GPUs.
The training iteration interval $\Delta T$ is set to 100 and the number of permutation candidates $K$ is set to 100.
We use the stochastic gradient descent (SGD) optimizer to perform sparse training. In the first 5 training epochs, the learning rate linearly increases from 0 to 0.1 and then is decayed using the cosine annealing~\cite{loshchilov2016sgdr}.
The momentum and batch size are respectively set to 0.9 and 256.
On CIFAR-10, we train all networks for 300 epochs with a weight decay of 1 $\times 10^{-3}$.
On ImageNet, we follow~\cite{zhou2021learning} to train ResNet-18/50 for a total of 120 epochs.
For DeiT-small, we follow~\cite{zhang2022learning} to train for 300 epochs in total using the timm framework~\cite{rw2019timm}.

\subsection{Comparison on CIFAR-10}\label{cifar10}
\begin{table}[t]
	\centering
	\caption{Comparison between different methods for training the N:M sparse ResNet-32 on CIFAR-10.}
\resizebox{1.0\columnwidth}{!}{
	\begin{tabular}[b]{ lcccccc}
		\toprule
		 Method & N:M  & Top-1 & Forward & Backward\\
		& Pattern & Accuracy (\%) & Acceleration & Acceleration \\
		\midrule
		 Baseline & - & 94.52  &\XSolidBrush  & \XSolidBrush  \\
		 SR-STE  &  2:4   &    94.68  &\Checkmark   & \XSolidBrush  \\
          T-Mask  &  2:4   &    94.52   &\Checkmark  &\Checkmark    \\
      \rowcolor[gray]{0.9}  Bi-Mask  &  2:4   &   \bf  94.78   &\Checkmark  &\Checkmark    \\
        \midrule
         SR-STE  &  1:4   &  \bf  94.52  &\Checkmark   & \XSolidBrush  \\
          T-Mask  &  1:4   &    94.04   &\Checkmark  &\Checkmark    \\
      \rowcolor[gray]{0.9}  Bi-Mask  &  1:4   &    94.43   &\Checkmark  &\Checkmark    \\
        \midrule
         SR-STE  &  1:16   &   \bf 92.92  &\Checkmark   & \XSolidBrush  \\
          T-Mask  &  1:16   &    92.02   &\Checkmark  &\Checkmark    \\
     \rowcolor[gray]{0.9}   Bi-Mask  &  1:16   &    92.77   &\Checkmark  &\Checkmark    \\
        \bottomrule
	\end{tabular}}
    \label{tab:r32_cifar10}
  \vspace{-0.6cm}
\end{table}

\textbf{ResNet-32.} 
We first apply our Bi-Mask to train ResNet-32 model.
The quantitative results are reported in Table\,\ref{tab:r32_cifar10}.
We can see from the table that the proposed Bi-Mask yields significantly better performance than the transposable mask at all N:M cases, and achieves comparable performance with the vanilla N:M mask that fails to achieve backward acceleration.
For example, Bi-Mask obtains 94.78\% Top-1 accuracy at 2:4 sparse pattern, surpassing SR-STE and T-Mask by 0.10\% and 0.26\%.
Therefore, these accuracy results well demonstrate the efficacy of our Bi-Mask. 

\begin{table}[t]
	\centering
	\caption{Comparison between different methods for training the N:M sparse MobileNet-V2 on CIFAR-10.}
 \resizebox{1.0\columnwidth}{!}{
	\begin{tabular}[b]{ lcccccc}
		\toprule
		 Method & N:M  & Top-1 & Forward & Backward\\
		& Pattern & Accuracy (\%) & Acceleration & Acceleration \\
		\midrule
		 Baseline & - & 94.43  &\XSolidBrush  & \XSolidBrush  \\
		 SR-STE  &  2:4   &    94.26  &\Checkmark   & \XSolidBrush  \\
          T-Mask  &  2:4   &   94.12    &\Checkmark  &\Checkmark    \\
     \rowcolor[gray]{0.9}   Bi-Mask  &  2:4   &    \bf 94.46   &\Checkmark  &\Checkmark    \\
        \midrule
         SR-STE  &  1:4   &   \bf 94.48  &\Checkmark   & \XSolidBrush  \\
          T-Mask  &  1:4   &   93.81   &\Checkmark  &\Checkmark    \\
      \rowcolor[gray]{0.9}  Bi-Mask  &  1:4   &     94.28  &\Checkmark  &\Checkmark    \\
        \midrule
         SR-STE  &  1:16   &  \bf  93.14  &\Checkmark   & \XSolidBrush  \\
          T-Mask  &  1:16   &    90.12  &\Checkmark  &\Checkmark    \\
      \rowcolor[gray]{0.9}  Bi-Mask  &  1:16   &   92.48   &\Checkmark  &\Checkmark    \\
        \bottomrule
	\end{tabular}}
    \label{tab:mobv2_cifar10}
\vspace{-0.5cm}
\end{table}

\textbf{MobileNet-V2.}
We further investigate the effectiveness of Bi-Mask for training N:M sparse MobileNet-V2, a prevailing network with a compact design of depth-wise separable convolution.
Table\,\ref{tab:mobv2_cifar10} again suggests a significantly higher accuracy of Bi-Mask compared with T-Mask under the same backward acceleration.
For instance, Bi-Mask maintains the performance of the dense model at 1:4 pattern, while T-Mask suffers apparent accuracy drops (94.28\%, 94.43\%, and 93.81\%) for Bi-Mask, dense model, and T-Mask, respectively).
%

%
\subsection{Comparison on ImageNet}\label{imagenet}

\textbf{ResNet-18.}
Table\,\ref{tab:r18_imagenet} shows the performance comparison of different methods for training 2:4 sparse ResNet-18 on ImageNet.
Compared with T-Mask~\cite{hubara2021accelerated}, the proposed Bi-Mask achieves 1.6\% performance gains.
Notably, Bi-Mask even surpasses ASP~\cite{nvidia2020a100} by 0.9\%, later of which fails backward acceleration. 
Therefore, the superiority of our proposed Bi-Mask on the large-scale dataset is validated.

\begin{table}[!t]
	\centering
	\caption{Comparison between different methods for training the N:M sparse ResNet-18 on ImageNet.}
 \resizebox{1.0\columnwidth}{!}{
	\begin{tabular}[b]{ lcccccc}
		\toprule
		 Method & N:M  & Top-1 & Forward & Backward\\
		& Pattern & Accuracy (\%) & Acceleration & Acceleration \\
		\midrule
		 Baseline & - & 70.9  &\XSolidBrush  & \XSolidBrush  \\
		ASP & 2:4 &69.9 &\Checkmark   & \XSolidBrush \\ 
		 SR-STE  &  2:4   &   71.2  &\Checkmark   & \XSolidBrush  \\
          T-Mask  &  2:4   &    69.2   &\Checkmark  &\Checkmark    \\
      \rowcolor[gray]{0.9}  \bf Bi-Mask  &  2:4   &   \bf  70.8   &\Checkmark  &\Checkmark    \\
        \bottomrule
	\end{tabular}}
    \label{tab:r18_imagenet}
  \vspace{-0.3cm}
\end{table}

\begin{table}[!t]
	\centering
	\caption{Comparison between different methods for training the N:M sparse ResNet-50 on ImageNet.}
        \resizebox{1.0\columnwidth}{!}{
	\begin{tabular}[b]{lcccccc}
		\toprule
		 Method & N:M  & Top-1 & Forward & Backward\\
		& Pattern & Accuracy (\%) & Acceleration & Acceleration \\
		\midrule
		 Baseline & - & 77.1  &\XSolidBrush  & \XSolidBrush  \\
		ASP & 2:4 &76.8 &\Checkmark   & \XSolidBrush  \\ 
		 SR-STE  &  2:4   &    77.0  &\Checkmark   & \XSolidBrush   \\
          T-Mask  &  2:4   &     76.2  &\Checkmark  &\Checkmark     \\
        \rowcolor[gray]{0.9}\bf Bi-Mask  &  2:4   &    \bf 77.4  &\Checkmark  &\Checkmark  \\
		\midrule
		 Baseline & - & 77.1  &\XSolidBrush  & \XSolidBrush  \\
		 SR-STE  &  4:8   &    77.4  &\Checkmark   & \XSolidBrush  \\
          T-Mask  &  4:8   &     77.1  &\Checkmark  &\Checkmark \\
        \rowcolor[gray]{0.9}\bf Bi-Mask  &  4:8   &    \bf 77.5 &\Checkmark  &\Checkmark \\
        \midrule
		 Baseline & - & 77.1  &\XSolidBrush  & \XSolidBrush  \\
		 SR-STE  &  1:4   &    75.3  &\Checkmark   & \XSolidBrush \\
          T-Mask  &  1:4   &    73.8  &\Checkmark  &\Checkmark \\
       \rowcolor[gray]{0.9}  \bf Bi-Mask  &  1:4   &     \bf 75.6 &\Checkmark  &\Checkmark \\
        \midrule
		 Baseline & - & 77.1  &\XSolidBrush  & \XSolidBrush \\
		 SR-STE  &  2:8   &    76.2 &\Checkmark   & \XSolidBrush  \\
          T-Mask  &  2:8   &     73.6  &\Checkmark  &\Checkmark \\
        \rowcolor[gray]{0.9}\bf Bi-Mask  &  2:8   &      \bf 76.3 &\Checkmark  &\Checkmark \\
                \midrule
		 Baseline & - & 77.1 &\XSolidBrush  & \XSolidBrush \\
		 SR-STE  &  1:16   &    71.5 &\Checkmark   & \XSolidBrush   \\
          T-Mask  &  1:16   &     66.4  &\Checkmark  &\Checkmark\\
       \rowcolor[gray]{0.9} \bf Bi-Mask  &  1:16   &   \bf 71.5 &\Checkmark  &\Checkmark \\
        \bottomrule
	\end{tabular}}
    \label{tab:r50_imagenet}
  \vspace{-0.3cm}
\end{table}

\textbf{ResNet-50.}
We further show the performance of training N:M sparse ResNet-50 on ImageNet.
As shown in Table\,\ref{tab:r50_imagenet}, Bi-Mask beats all the competitors across all N:M cases with the same or superior acceleration results.
In particular, in comparison with SR-STE that gets stuck in dense backward propagation, Bi-Mask results in backward acceleration, meanwhile shows the best performance.
For example, it surpasses SR-STE by 0.3\% at 1:4.
As for T-Mask that also accelerates the backward propagation, our T-Mask shows superior performance in particular to the cases with a smaller N value. As analyzed in Sec.\,\ref{trans}, a small N or M greatly degrades the mask flexibility of T-Mask, therefore severe performance drops occur.

Next, we compare the efficiency between searching the optimal N:M transposable mask~\cite{hubara2021accelerated} and our permutation for the backward mask.
We report the runtime for searching ResNet-50 with different N:M cases on one NVIDIA Tesla A100 GPU.
Table\,\ref{tab:r50_speed} suggests superior efficiency of our Bi-Mask. For example, it takes negligible 15.0s for our Bi-Mask to find a good permutation at 1:16. As a contrast, T-Mask requires around 278.4s under the same N:M setting. 
Given the efficiency and accuracy, the efficacy of Bi-Mask is evident.


%
\begin{table}[t]
	\centering
	\caption{Time comparison (s) between T-Mask and Bi-Mask for searching N:M masks of ResNet-50 at different patterns.}
 \resizebox{0.8\columnwidth}{!}{
	\begin{tabular}[b]{ lcccccc}
		\toprule
		 Method & 2:4  & 4:8 & 1:4 & 2:8 & 1:16\\
		\midrule
          T-Mask  &  155.1 & 193.2 & 168.6 & 200.1 & 278.4  \\
    \rowcolor[gray]{0.9}   \bf Bi-Mask   &  15.5 & 14.3 & 13.6 & 15.7 & 15.0   \\
        \bottomrule
	\end{tabular}}
    \label{tab:r50_speed}
     \vspace{-0.2cm}
\end{table}

\begin{table}[t]
	\centering
	\caption{Comparison between different methods for training the N:M sparse DeiT-small on ImageNet.}
\resizebox{1.0\columnwidth}{!}{
	\begin{tabular}[b]{ lcccccc}
		\toprule
		 Method & N:M  & Top-1 & Forward & Backward\\
		& Pattern & Accuracy (\%) & Acceleration & Acceleration \\
		\midrule
		 Baseline & - & 79.8  &\XSolidBrush  & \XSolidBrush  \\
		 SR-STE  &  2:4   &    75.7  &\Checkmark   & \XSolidBrush   \\
          T-Mask  &  2:4   &    71.5  & \Checkmark  & \Checkmark  \\
      \rowcolor[gray]{0.9} \bf Bi-Mask  &  2:4   &  \bf 77.6 & \Checkmark   & \Checkmark  \\
        \bottomrule
	\end{tabular}}
    \label{tab:deit_imagenet}
       \vspace{-0.5cm}
\end{table}

\textbf{DeiT-small.}
We further continue to compare the efficacy of Bi-Mask for training 2:4 sparse DeiT, an advanced vision transformer model.
As manifested in Table\,\ref{tab:deit_imagenet}, the proposed Bi-Mask consistently obtains the best Top-1 under different N:M cases, with its additional merit in backward acceleration for N:M sparse training.
For instance, Bi-Mask improves the Top-1 accuracy of SR-STE by 1.7\% at 2:4, and gains 5.9\% Top-1 improvements over off-the-shelf T-Mask.
The above observations well demonstrate the ability of Bi-Mask in compressing and accelerating the recent advanced vision transformer models.

\subsection{Performance Study}\label{sec:ablation}
In this section, we investigate the performance of Bi-Mask to respectively explore the effectiveness of its components. All the experimental results are conducted on ImageNet using ResNet-18.

\begin{figure}[!t]
    \centering
    \includegraphics[width=0.45\textwidth]{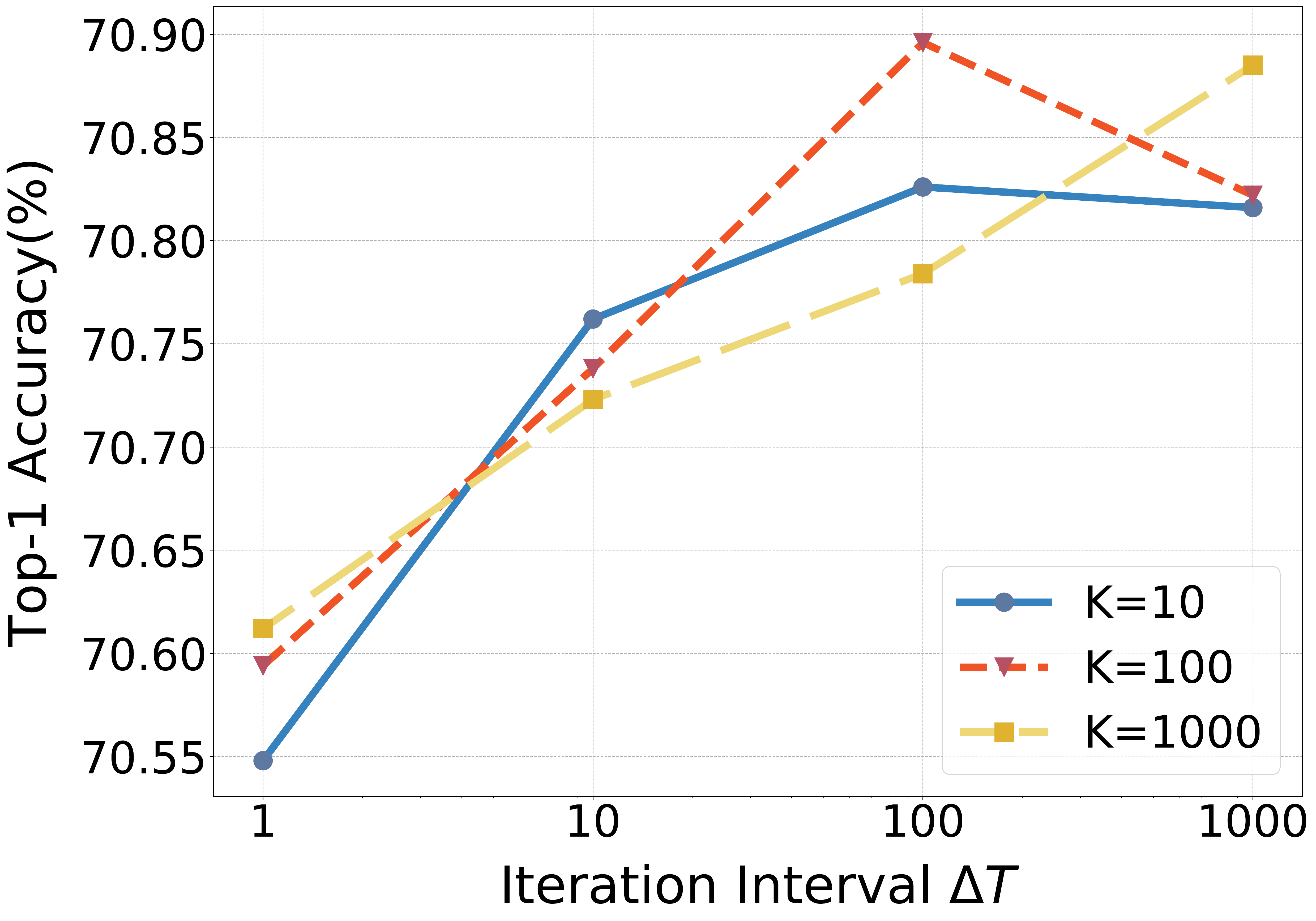}
    \caption{Performance influence of the training iteration inverval $\Delta T$ and candidation number $K$ in our permutation updating.}
    \label{fig:permutation_search}
\end{figure}

\textbf{Permutation Updating.} We first perform ablation studies of the permutation updating. 
In Fig.\,\ref{fig:permutation_search}, we examine the performance of Bi-Mask under different training iteration intervals $\Delta T \in [1, 10, 100, 1000]$ and permutation candidate number $K \in [10, 100, 1000]$.
As can be observed, the best accuracy is obtained when the permutation updating is performed every 100 training iterations.
To analyze, small intervals incurs unstable sparse training as the typology of computing graph change in an excessive frequency.
In contrast, large intervals outdate the optimal permutation, thus also leading to worse performance.
Besides, the performance becomes saturated when the candidate number reaches 100 or more.
The result shows that it is unnecessary to construct a time-consuming greedy algorithm to find out the optimal permutation.

\textbf{Binarization Criteria.} We further investigate the binarization criteria used to force the transposed weights to be N:M sparse after the permutation.
Recall we opt magnitudes of sparse weights in Eq.\,(\ref{eq:backward_mask_final}) for a better fit with the forward mask to reduce the gradient gap.
For comparison, we consider three variants including 1) preserving weights with the Top-N largest magnitudes of their gradients; 2) sampling N weights from the multinomial probability distribution according to their magnitudes; 3) randomly preserving N weights.
Table\,\ref{tab:ablation_criteria} manifests our experimental results for the comparison.
%
%
We can observe that the variants reduce the accuracy more or less. Among them, random one incurs the most performance drops since it severely enlarges the dissimilarity between the forward and backward masks and causes great gradient gaps. As for our weight magnitude, it well complies with the forward mask settings, therefore a better result can be observed.

\textbf{Ablation Study.} To further understand the effect of each component in our Bi-Mask, we conduct an ablation study by starting from our training baseline of the vanilla N:M mask~\cite{zhou2021learning} (denoted by Baseline), and gradually including the backward mask and permutation updating.
Table\,\ref{tab:ablation_search} shows that our backward mask enables backward acceleration with a Top-1 accuracy drop of 0.3\%. As an analysis, the little degradation mainly yields from that our backward mask sometimes mistakenly eliminate gradients of some non-zero masked weights, as discussed in Sec.\,\ref{sec:bi-mask}.
After further adding our proposed permutation updating, the performance of sparse ResNet-18 even increases by 0.3\% on the basis of the Baseline method. This is because our permutation updating operation results in more eligible N:M blocks and reduce the possibility of incorrect gradient elimination.
In conclusion, each part of our proposed Bi-Mask in this paper plays a unique role in boosting the performance of our N:M sparse training.

\begin{table}[!t]
\small
    \caption{Performance of different binarization criteria for backward mask in Bi-Mask.
    }
    \centering
    \resizebox{1.0\columnwidth}{!}{
    \begin{tabular}{cccc>{\columncolor[gray]{0.9}}c}
    \toprule
Model & Criteria & Pattern& Top-1 Accuracy(\%) \\
   \midrule
ResNet-18 & Gradient Magnitude  &2:4& 70.56\\
ResNet-18 &  Multinomial Sampling & 2:4 & 70.42\\
 ResNet-18 & Random &2:4 & 67.76\\
 \rowcolor[gray]{0.9} ResNet-18 & Weight Magnitude  & 2:4&  70.73\\
    \bottomrule
    \end{tabular}}
    \label{tab:ablation_criteria}
\end{table}

\begin{table}[!t]
\small
    \caption{Ablation study for the proposed Bi-Mask.
    }
    \centering
    \resizebox{1.0\columnwidth}{!}{
    \begin{tabular}{cccc>{\columncolor[gray]{0.9}}c}
    \toprule
Model & Criteria & Pattern& Top-1 Accuracy(\%) \\
   \midrule
   ResNet-18 & Baseline  &2:4& 70.5\\
   ResNet-18 & + Backward Mask  &2:4& 70.2\\
ResNet-18 & + Permutation Updating  & 2:4&  70.8\\
    \bottomrule
    \end{tabular}}
    \label{tab:ablation_search}
\end{table}

\section{Limitations}
In this section, we further discuss unexplored limitations of Bi-Mask in this paper, which will be our major future focuses. First, following the
compared methods, we only train N:M sparse networks on the image classification task. More efforts can be made to verify the effectiveness of Bi-Mask on other tasks such as object detection.
Besides, we only explore the acceleration on the forward and backward propagation, while accelerating the update phase of weights~\cite{chmiel2022optimal} remains to be excavated in the near future.
At last, our random generation for the permutation does
not always guarantee to maximize the number of N:M blocks. Though it does not damage the overall performance, a better solution is expected to be explored for possibly locating the best permutation.

\section{Conclusion}
In this work, we have presented a novel Bi-direction Mask (Bi-Mask) for efficient N:M find-grained sparse neural network training.
Instead of imposing a transposable constraint on the N:M sparse mask, our Bi-Mask independently builds masks in the forward and backward directions.
The mask in the backward direction is obtained through an efficient permutation in the weight rows and a following magnitude-based pruning to enable acceleration on the N:M sparse tensor core.
Extensive experiments have demonstrated the superiority of Bi-Mask over several SOTAs.
Particularly, Bi-Mask surpasses its competitor by a large margin under the same acceleration effects, and can even perform on par or even better than off-the-shelf methods that often fail to achieve backward acceleration.
%
%

\section*{Acknowledgement}
This work is supported by the National Science Fund for Distinguished Young (No.62025603), the National Natural Science Foundation of China (No.62025603, No. U1705262, No. 62072386, No. 62072387, No. 62072389, No, 62002305, No.61772443, No. 61802324 and No. 61702136) and Guangdong Basic and Applied Basic Research Foundation (No.2019B1515120049).


\bibliography{egbib}
\bibliographystyle{icml2023}

\end{document}